%% file: acl2021.tex
\documentclass[11pt,a4paper]{article}
\usepackage[hyperref]{acl2021}
\usepackage{times}
\usepackage{latexsym}

\usepackage{microtype}
\usepackage{booktabs}

\aclfinalcopy 

\usepackage{amsfonts}
\usepackage{amssymb}
\usepackage{amsthm}
\usepackage{float}
\usepackage{enumerate}
\usepackage{booktabs}
\theoremstyle{plain}

\usepackage{adjustbox}
\usepackage{xfrac}
\usepackage{breqn}
\usepackage{scalerel}
\usepackage{subfig}
\usepackage{xspace}
\usepackage{url}
\usepackage{graphicx}
\usepackage{pgfplots}
\pgfplotsset{
tick label style={font=\small},
label style={font=\small},
title style={font=\normalsize},
legend style={font=\footnotesize}
}
\usepackage[most]{tcolorbox}

\theoremstyle{definition}

\usepackage{times}
\usepackage{latexsym}

\usepackage{emoji}
\newcommand{\ucambridge}{\emoji[twitter]{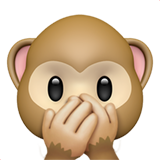}}
\newcommand{\ethz}{\emoji[twitter]{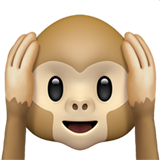}}
\newcommand{\google}{\emoji[twitter]{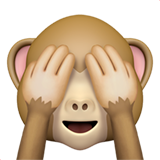}}

\usepackage[T1]{fontenc}

\usepackage[utf8]{inputenc}

\usepackage{microtype}

\usepackage{cleveref}
\crefname{section}{\S}{\S\S}
\Crefname{section}{\S}{\S\S}
\crefname{table}{Table}{Tables}
\crefname{figure}{Figure}{Figures}
\crefname{algorithm}{Algorithm}{}
\crefname{equation}{eq.}{}
\crefname{appendix}{App.}{}
\crefformat{section}{\S#2#1#3}

\usepackage{float}
\newtheorem*{ex1}{\textit{Ex.~A}}
\newtheorem*{ex2}{\textit{Ex.~B}}

\usepackage{todonotes}
\makeatletter
\newcommand*\iftodonotes{\if@todonotes@disabled\expandafter\@secondoftwo\else\expandafter\@firstoftwo\fi} 
\makeatother

\newcommand{\yy}{\mathbf{y}}
\newcommand{\ww}{\mathbf{w}}

\newcommand{\vtheta}{{\boldsymbol \theta}}
\newcommand{\ptheta}{p_{\scaleto{\vtheta}{4pt}}}

\newcommand{\vocab}{\mathcal{V}}

\usepackage{tipa}

\title{A Cognitive Regularizer for Language Modeling}

\author{
{Jason Wei\raise1.0ex\hbox{\normalfont\google}\raise1.0ex\hbox{\normalfont}}~~~~\;~Clara Meister\raise1.0ex\hbox{\normalfont\ethz}~~~~\;~Ryan Cotterell\raise1.0ex\hbox{\normalfont\ethz,\ucambridge}
\\
  \raise1.0ex\hbox{\normalfont\google}Google AI Language~~~\;~\raise1.0ex\hbox{\normalfont\ethz}ETH Z{\"u}rich~~~\;~\raise1.0ex\hbox{\normalfont\ucambridge}University of Cambridge \\
  \texttt{jasonwei@google.com}~~~~\;~\texttt{clara.meister@inf.ethz.ch} \\ \texttt{ryan.cotterell@inf.ethz.ch}
}

\date{}

\begin{document}
\maketitle
\begin{abstract}
The uniform information density (UID) hypothesis, which posits that speakers behaving optimally tend to distribute information uniformly across a linguistic signal, has gained traction in psycholinguistics as an explanation for certain syntactic, morphological, and prosodic choices.
In this work, we explore whether the UID hypothesis can be operationalized as an inductive bias for statistical language modeling. 
Specifically, we augment the canonical MLE objective for training language models with a regularizer that encodes UID. In experiments on ten languages spanning five language families, we find that using UID regularization consistently improves perplexity in language models, having a larger effect when training data is limited.
Moreover, via an analysis of generated sequences, we find that UID-regularized language models have other desirable properties, e.g., they generate text that is more lexically diverse. Our results not only suggest that UID is a reasonable inductive bias for language modeling, but also provide an alternative validation of the UID hypothesis using modern-day NLP tools.\looseness=-1
\end{abstract}


\section{Introduction}
Language has been hypothesized to follow certain information-theoretic constraints. 
One of the most famous of these constraints is the uniform information density (UID) hypothesis \cite{fenk1980konstanz,jaeger2010redundancy}, which states that, subject to the rules of the grammar, speakers aim to distribute information density across a linguistic signal as uniformly as possible.
That is, speakers behaving optimally should structure their utterances such that the differences between the peaks and troughs in information are minimized.

In the psycholinguistics literature, the UID hypothesis has been used to explain a variety of linguistic phenomena ranging from how we shorten the phonetic duration of more-predictable linguistic  
\begin{figure}[H]
    \centering
    (a)
    \includegraphics[width=0.82\linewidth]{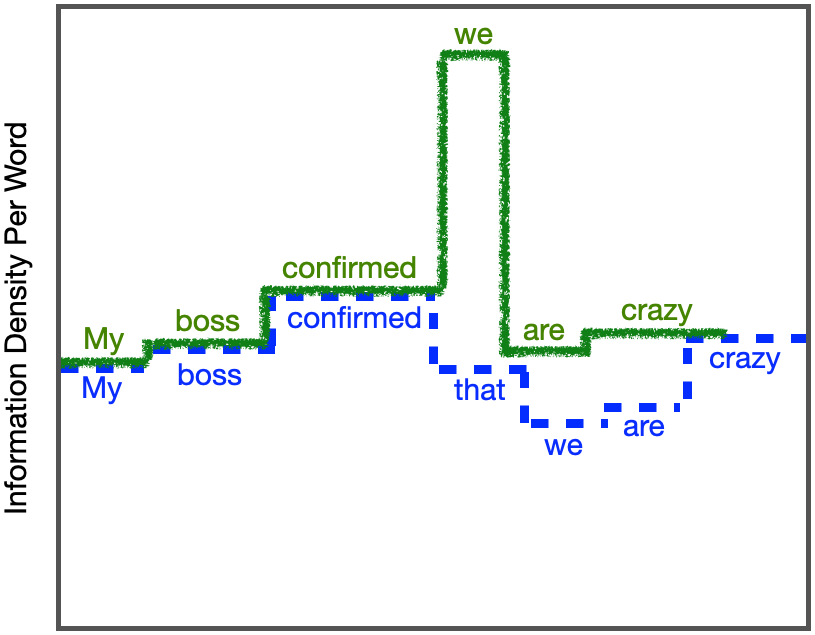}\\
    (b)
    \includegraphics[width=0.82\linewidth]{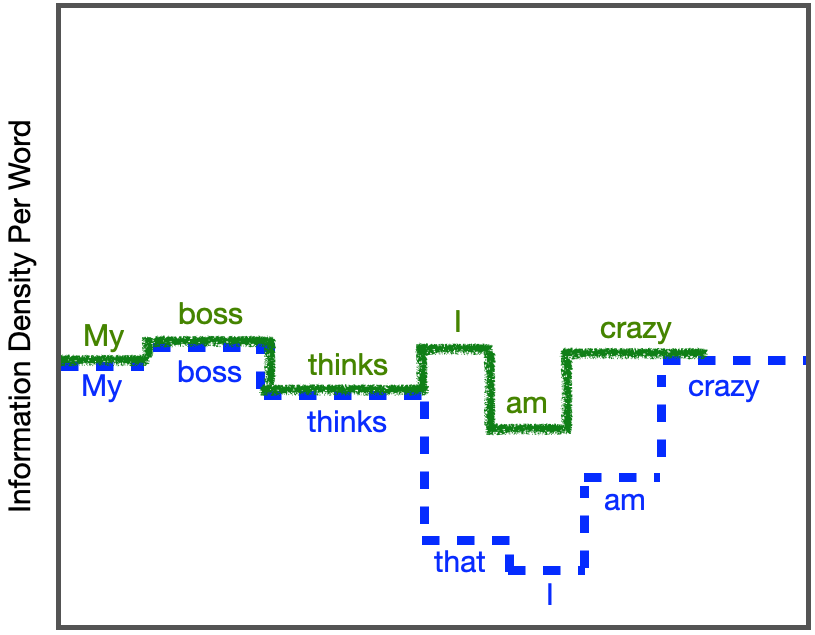}
    \caption{
    Graphical illustration of two examples regarding UID.
    In (a), many speakers will prefer the version with the relativizer \textit{that} (dotted blue line). 
    The UID hypothesis posits that this is because, without the relativizer, the first word of the relative clause, \textit{we}, has high information density; and so including the relativizer distributes the per-word information density more uniformly. 
    In (b), the relativizer \textit{that} is often omitted because, at the onset of the relative clause, the information density of \textit{I} is lower and therefore the distribution of information density is already relatively uniform.
    Illustration based on \citet{jaeger2010redundancy}.
    }
    \label{fig:uid-example}
    \vspace{-2mm}
\end{figure}
\noindent units \cite{aylettturk04} to when we decide to use optional syntactic relativizers \cite{jaeger-levy-07}, among other phenomena \cite{bell2003effects,frank2008speaking}. 
These studies often use language models to estimate the information density of linguistic units, taking observations of low variation of information density in well-formed utterances as evidence for the UID hypothesis.

In this paper, we propose a new experimental paradigm that uses modern-day NLP models to test the UID hypothesis. 
Whereas prior work has used language modeling as a tool for observing UID,\footnote{On its own, the term `UID' is formally an \emph{attribute} of a linguistic signal. We also use it throughout this work to refer to the \emph{concept} that UID is a desirable property.} we explore the converse---can UID be used as a tool to train better language models?
Specifically, if the UID hypothesis is true, then we should be able to operationalize UID as a regularizer to help train language models.
Moreover, observing lower perplexity in language models trained with this regularization would imply that the concept of UID is a good inductive bias for language modeling, thereby providing a new type of evidence for the UID hypothesis \emph{at scale}.\looseness=-1

In experiments, we indeed find such evidence: across a variety of languages and dataset sizes, UID regularization consistently improves performance, having a larger effect when training data is limited. 
Moreover, we observe that---in comparison with their unregularized counterparts---UID-regularized language models are (1) higher entropy while achieving the same (or better) test set perplexity and (2) generate text that is longer and more lexically diverse. 
Our work is the first to explore the interaction between UID and \emph{training} modern-day neural language models, and our findings---that a cognitively motivated objective can improve language model performance---open up new avenues for testing other psycholinguistic hypotheses in a similar framework.

\section{Preliminaries: Language Modeling} \label{sec:preliminaries}
The task of language modeling aims to estimate a model of the probability of observing any given string in (a subset of) natural language. 
Formally, a language model $p$ is an (unconditional) probability distribution over sequences of words $\ww = \langle w_1, w_2,\dots \rangle$, where $\ww$ consists of tokens from some vocabulary and begins and ends with special tokens \textsc{bos} and \textsc{eos}, respectively.

Today's language models are typically parameterized by neural networks (e.g., transformers \cite{vaswani_att}), that follow a local-normalization scheme. Specifically, the model provides a conditional distribution over the vocabulary at each time step; we can then compute the probability of an entire sequence $\ww$ as: 
\begin{equation}
     \ptheta(\ww) = \prod_{t=1}^{|\ww|} \ptheta(w_t \mid \ww_{<t})
\end{equation}
\noindent where $\vtheta$ are the parameters of the model and we use $\ww_{<t}$ to represent the first $t-1$ tokens of $\ww$.
Parameters are estimated by optimizing over some objective $L(\vtheta)$. The standard objective for language modeling is the negative log-likelihood of a dataset $\mathcal{W}$ under the model:
\begin{align} \label{eq:lm_nll}
L(\vtheta) &= - \sum_{\ww \in \mathcal{W}} \log \ptheta(\ww)
\end{align}

\noindent Subsequently, we drop explicit dependence on $\vtheta$ when it is obvious from context. 

To assess the goodness of fit of a model $p$, we typically evaluate its perplexity on some held-out dataset $\mathcal{W}_{\mathrm{test}}$, where perplexity (PPL) is defined as 
\begin{equation}
    \mathrm{PPL}(p) = \exp\left(-\sum_{\ww \in \mathcal{W}_{\mathrm{test}}} \frac{1}{|\ww|} \log p(\ww)\right)
\end{equation}
\noindent Note that under this definition of perplexity, our evaluation metric is slightly different than the training objective; the former computes an average over each sequence while the later treats all tokens equally, regardless of the length of the sequence in which they are present.

\section{Uniform Information Density}
Communication via natural language is a complicated and nuanced process that takes place under a host of cognitive and environmental constraints. 
As a result, speakers have to make (perhaps subconscious) choices to best navigate this communicative dance. A rational speaker would use these choices to optimize the communicative properties of their utterances. One such locus of optimization is outlined by the \textbf{Uniform Information Density (UID)} hypothesis. 

\subsection{The UID Hypothesis}
At its core, the UID hypothesis aims to explain certain phenomena in human language processing using an information-theoretic approach: we can view language as a transfer of information, which is transmitted with a certain density through a communication channel. The UID hypothesis posits that speakers that behave optimally will structure their utterances to avoid peaks and troughs in this information density \cite{aylettturk04,jaeger-levy-07,jaeger2010redundancy}. 
More formally stated: 
``\emph{Within the bounds defined by grammar, speakers prefer utterances that distribute information uniformly across the signal (information density). Where speakers have a choice between several variants to encode their message, they prefer the variant with more-uniform information density (ceteris paribus)}'' \cite{jaeger2010redundancy}.

\subsection{Example: UID in syntactic reduction}
To better understand the UID hypothesis, consider the concrete example of syntactic reduction (\textit{that}-mentioning) from  \citet{jaeger2010redundancy}, which we show graphically in \cref{fig:uid-example} and also describe below.

\begin{ex1}
My boss confirmed [that] we are crazy.
\end{ex1}
\begin{ex2}
My boss thinks [that] I am crazy.
\end{ex2}

In both these sentences, the use of the relativizer \textit{that} is syntactically optional---at the onset of a relative clause (RC), speakers can, but do not have to, include the relativizer. 
Many speakers, however, would argue that the sentence flows better with the relativizer included in Example A and the relativizer omitted in Example B.

The UID hypothesis provides a potential explanation for this phenomenon.
When a RC is used without a relativizer, the first word of the RC conveys two pieces of information: both the onset of the RC, as well as part of the RC's internal contents. 
In Example A, many speakers would find that the information density of the first word in the RC, \textit{we}, is high, and so adding in the relative clause distributes the information over two words, making it easier to parse.
In Example B, the information density of the first word in the RC, \textit{I}, is lower relatively, and so we do not need to (or it is not as beneficial to) include the relativizer.

\subsection{Measuring UID}
Now that we better understand what the UID hypothesis attempts to explain, how might we operationalize UID and find quantitative evidence of the pressure for it in language?
First, to quantify the amount of information conveyed by a word, we turn to the most basic information-theoretic definition: the information conveyed by a word $w$ in context is its Shannon information content \cite{shannon1948mathematical}, also called \textit{surprisal}.
Ideally, this surprisal would be measured using the ``true'' distribution over human language. Because we do not have access to such a distribution, we often estimate it using a statistical language model.
That is, given a statistical language model $p$, which estimates the probability of a word given its context, the surprisal $u(w_t)$ of word $w_t$ is defined as the following: 

\begin{equation}\label{eq:surprisal}
    u(w_t) = -\log p(w_t \mid \textbf{w}_{<t})
\end{equation}
\noindent This setup provides a natural approach to exploring how UID might manifest---if the UID hypothesis is true, then we should observe that variation in surprisal, as estimated by a language model, is minimized in natural language.

Using this approach, prior work has accumulated evidence for UID across various levels of linguistic representation \cite[][\textit{inter alia}]{pluymaekers2005lexical,bell2009predictability}.
As some of the earliest examples, \citet{aylettturk04} showed that linguistic units that had high surprisal according to a tri-gram language model were uttered with longer syllable durations, and \citet{jaeger-levy-07} found that for RCs in which the first word had higher surprisal, relativizers were more likely to be used in the RC during actual speech. Further examples are given in our related work section (\cref{sec:discussion-related-work}).


\section{UID-Regularized Language Modeling}

While prior work has shown evidence that UID can help explain many of the choices we make when generating language, to the best of our knowledge, operationalizations of UID have not been explicitly employed as part of the training objective in modern-day NLP models. 
This raises the simple question that is central to our paper:

\begin{quote}
\centering
\begin{tcolorbox}[colback=blue!5!white,colframe=blue!75!black]
\textit{Can UID serve as an inductive bias for training statistical language models?}
\end{tcolorbox}
\end{quote}

In an effort to answer this question, we present a scheme for incorporating operationalizations of UID into the language model training objective. Formally, we augment the canonical maximum likelihood estimation objective\footnote{Note that the maximum likelihood estimation objective minimizes (over $\ww \in \mathcal{W}$) $-\log p(w_t \mid \textbf{w}_{<t})$, i.e., surprisal. Although such an objective may indirectly minimize peaks and dips in surprisal across a sequence simply by pushing them towards 0, it does not explicitly include any sequence level penalty for even surprisal distribution.\looseness=-1} in \cref{eq:lm_nll} with UID operationalizations as regularizers $\mathcal{R}$.
Under this new objective, we minimize
\begin{equation}\label{eq:reg_obj}
    L_{\mathcal{R}}(\vtheta ) =  L(\vtheta ) + \beta \cdot \mathcal{R}(\vtheta) 
\end{equation}
where $\beta > 0$ is the strength coefficient of the regularizer.
We consider two natural operationalizations of UID---inspired by \citet{collins2014information}---as regularizers for training language models: 

\paragraph{Variance Regularizer.}
UID concerns the distribution of information in language production, and so a natural measure of this behavior is the variance of surprisals.
Thus, we first consider a regularizer that penalizes high variance among the surprisals of words in a given sequence:
\begin{equation}\label{eq:var_reg}
    \mathcal{R}(\vtheta) = \frac{1}{|\ww|} \sum_{t=1}^{|\ww|} (u(w_t) - \mu)^2 
\end{equation}
where $\mu = \frac{1}{|\ww|}\sum_{t=1}^{|\ww|} u(w_t)$. Note that here, and in our subsequent regularizers, we estimate $u(\cdot)$ via \cref{eq:surprisal} using our model $\ptheta$.
\paragraph{Local Consistency.}
Next, we consider a local consistency regularizer that encourages the surprisals of adjacent words to have similar magnitude:
\begin{equation}\label{eq:lc_reg}
    \mathcal{R}(\vtheta) = \frac{1}{|\ww| - 1} \sum_{t = 1}^{|\ww|-1} \Big(u(w_t) - u(w_{t+1})\Big)^2
\end{equation}
This regularizer is also a reasonable operationalization of UID---if every surprisal is similar to its neighbor, then the density of information in the sequence will be close to uniform.

Though we focus on these two regularizers, other operationalizations of UID certainly exist. 
For example, a similar variant of the above regularizers is the max regularizer \cite{meister-etal-2020-beam}, which penalizes the highest surprisal in a sentence.\footnote{We also tried this operationalization in preliminary experiments, but results were not as strong as the variance or local consistency regularizers.} 
Furthermore, UID may also be defined in terms of parse steps \cite{hale-2001-probabilistic2} or structural integrations \cite{gibson2000dependency}, as well as in spoken language in the form of filler words like \textit{uh} and \textit{um} or word repetition during challenging lexical retrieval. 
We consider these operationalizations (as well as the broader discussion of how to operationalize UID) as future work.

\section{Experimental Setup}
To empirically evaluate UID regularization, we train various language models with the UID-regularized objective (\cref{eq:reg_obj}) using the following experimental setup.

\paragraph{Datasets.}
We employ datasets from multiple languages and of varying sizes. 
We use the EuroParl corpus \cite{koehn2005europarl}---a multi-lingual dataset of discussions from the European Parliament that has been commonly used for language modeling \cite{cotterell-etal-2018-languages,mielke-etal-2019-kind}---since it is roughly semantically controlled in that all utterances are presumably about the same topics. 
We use EuroParl v7 download from the ACL 2014 SMT Workshop\footnote{\url{http://statmt.org/wmt14/translation-task.html}} and perform a 80--10--10 train-dev-test split on all five languages---Czech, English, French, German, and Spanish---which yields 46.7, 42.2, 47.2, 51.3, and 12.4 million training tokens for each language respectively.

Moreover, we experiment on languages from several language families; the five languages in Europarl that we consider are all Indo-European,
and so we look to Wiki-40B \cite{guo-etal-2020-wiki}, which contains Wikipedia dumps of a wide range of languages. 
We choose a set of diverse languages with training set sizes relatively similar to that of EuroParl: Finnish (a Uralic language; 59.3M training tokens), Indonesian (an Austronesian language; 45.7M training tokens), and Turkish (a Turkic language; 38.1M training tokens).
To explore performance on lower-resource languages, we additionally experiment with Swahili\footnote{Since there are no Niger-Congo languages in Wiki-40B, we perform a 80-10-10 split on Swahili Wikidumps (see \url{https://github.com/google-research/bert/blob/master/multilingual.md}).} (a Niger-Congo language; 6.3M training tokens) and Tagalog (an Austronesian language; 4.2M training tokens).
For all languages, we performed tokenization using the MosesTokenizer.\footnote{\url{https://pypi.org/project/mosestokenizer/}}
Train, dev, and test set splits are shown in \cref{tab:data_stats} in the Appendix.

\paragraph{Model Framework and Architecture.}
For our experiments, we use the \texttt{fairseq} library \cite{ott-etal-2019-fairseq}, a standard sequence modeling toolkit in PyTorch.
As our model, we use \texttt{fairseq}'s default transformer (with six decoder layers and eight attention heads), which achieves competitive\footnote{On Wikitext-103, the largest dataset we train on  (103 million tokens), we achieve a competitive perplexity of 29.89 (c.f. \citet{DBLP:journals/corr/abs-1803-08240}). For smaller datasets, we tried a smaller transformer architecture of four decoder layers and four attention heads, but it did not perform better than the six decoder layer and eight attention heads version, suggesting that this architecture was not too large for the datasets we use in this paper (even the Tagalog dataset we use is larger than the commonly used Penn Treebank and WikiText-2).} 
language modeling performance (although the purpose of our paper is not to achieve or compare with the state of the art).
For all experiments, we followed the data-preprocessing scripts and recommended hyperparameters provided in \texttt{fairseq}'s language modeling module; more detailed information can be found on the Github page.\footnote{\url{https://github.com/pytorch/fairseq/tree/master/examples/language_model}}

\paragraph{UID Regularizers.}
For UID regularization, we experiment with the variance (\cref{eq:var_reg}) and local consistency regularizers (\cref{eq:lc_reg}). 
We found in preliminary experiments that effective regularization strengths were often near $\beta=0.01$, and so we performed a grid search over values within an order of magnitude around $\beta=0.01$: $\beta \in \{0.006$, $0.008$, $0.01$, $0.02$, $0.03$, $0.04$, $0.05\}$. 
We choose the model with the lowest dev loss to evaluate on the test set.

\section{Results}

In this section, we report results for models trained under the UID-regularized objective.
We find that UID regularization consistently improves perplexity for models trained on various languages (\cref{subsec:multiple_languages}) and dataset sizes (\cref{subsec:dataset_size}).
Additionally, we examine properties of text generated by UID-regularized models (\cref{subsec:generated_text_evaluation}) and analyze the relationship between our operationalization of UID and perplexity (\cref{subsec:uid_behavior}).\looseness=-1

\subsection{Languages} \label{subsec:multiple_languages}
\cref{tab:multiple_languages} shows the results of UID-regularized language models trained on various languages from EuroParl and Wiki-40B, and includes statistical significance of changes in perplexity, as compared with baselines, computed using permutation tests\footnote{\url{http://www2.stat.duke.edu/~ar182/rr/examples-gallery/PermutationTest.html}} \cite{efron1994introduction}.
For all languages, UID regularization significantly improves perplexity for at least one of the two regularizers. 
Further- 
\input{tables/multiple_languages}
\hspace{-1.3mm}more, UID regularization (under the best performing $\beta$) never leads to worse perplexity.
These results suggest that incorporating UID operationalizations into a model's training objective leads to a better model of language, substantiating uniform information density as a valid inductive bias. 
Moreover, the improvement for many languages corroborates the expectation that UID should, due to its information theoretic nature, hold across languages \cite{jaeger_utility}.

\subsection{Dataset Size} \label{subsec:dataset_size}
Notably, we observe the largest improvements (1.6--2.9\%) in perplexity in \cref{tab:multiple_languages} for the lowest resource languages, Tagalog and Swahili (with 4.2 and 6.3 million training tokens respectively). 
Conversely, improvement was most marginal (0.2--0.5\%) on the highest-resource languages, French and Finnish (51.3 and 59.3 million training tokens respectively).
To remove language as a confounding factor from this observation, we perform a controlled analysis of the effects of UID regularization as a function of dataset size.

\input{tables/dataset_size_1}
We focus on English; in addition to the result on English EuroParl 2014 from \cref{tab:multiple_languages}, which contains 47.0 million training tokens, we experiment with the smaller monolingual English dataset from the 2006 NAACL Workshop on Statistical Machine Translation (WMT'06),\footnote{We downloaded the given train-dev-test splits from \url{https://www.statmt.org/wmt06/}.} which has 17.0M tokens in its training set, as well as the larger Wikitext-103 benchmark \cite{merity2016pointer}, which contains 103 million tokens in its training set.

\cref{tab:dataset_size_1} shows the perplexities for models with and without UID regulariztion for these three datasets.
As suggested by earlier results, improvements were strongest for the WMT'06 dataset, with an improvement of 1.4 perplexity points for the variance regularizer and 0.9 PPL points for local consistency. 
For the larger EuroParl and WT-103 datasets, on the other hand, improvement was more modest, ranging from 0.1 to 0.3 perplexity points. 

As further confirmation that UID regularization has a greater impact on smaller datasets, we perform an ablation study that roughly controls for language content by training models on the subsets of the same dataset.
For this ablation, we take subsets of 2, 4, 8, 12, 16, 24, and 32 million sentences from the 47 million sentences in English EuroParl, and observe how much the UID regularizers improve perplexity for each training dataset size.
As shown in \cref{fig:dataset_size}, the results tell the same story as \cref{tab:dataset_size_1}---UID regularization improves perplexity more for smaller datasets.
\input{figures/dataset_size_2}

These results are consistent with the expectation that models trained on smaller datasets are more likely to overfit and could therefore benefit more from regularization \cite{DBLP:conf/iclr/MelisDB18}.
As it is possible that the models trained on smaller datasets could benefit from any kind of regularization, we experiment with label smoothing \cite{szegedy2016rethinking}, another regularization technique that similarly augments the training objective with a penalty.
\cref{tab:label_smoothing} shows these results for models trained on WMT'06 and EuroParl with label smoothing---our experiments indicate that, across the board, label smoothing leads to worse perplexity compared with baseline models.\footnote{This negative result for applying label smoothing to language modeling is consistent with prior empirical findings \cite{muller2019does,gao-etal-2020-towards,meister+al.acl20}.} 
We take this result as further evidence that the improvement from UID regularization stems from the UID hypothesis as a valid inductive bias, rather than simply a need for any kind of regularization when training on smaller datasets.\looseness=-1
\input{tables/generated_text_evaluation}
\input{tables/label_smoothing}

\subsection{Evaluating Generated Text} \label{subsec:generated_text_evaluation}
Unconditional models of language have been observed to produce generic text that can be short, bland, or repetitive \cite{fan-etal-2018-hierarchical,kulikov-etal-2019-importance,curious-case}, and so in this subsection we investigate how UID regularization might affect these characteristics in generated text. 
For these experiments, we consider the baseline model, the variance-regularized model, and the local consistency-regularized model trained on English EuroParl.
To obtain text samples, we generate samples by sequentially sampling tokens according to the model's predicted distribution until the end-of-sequence (\textsc{eos}) token is sampled, i.e., ancestral sampling. Note that for language model $p$, this sampling scheme is equivalent to directly sampling $\yy \sim p$.
We obtain 10,000 samples for each model and report statistics in \cref{tab:generated_seq_analysis}.

We analyze each set of generated sentences for several metrics.
First, we compute the average length of generated sentences.
Next, we evaluate the lexical diversity of generated texts by computing the percent of unique $n$-grams for $n \in \{2, 3, 4\}$. 
Finally, sampling from a model also gives us a means for estimating the language model's entropy:
\begin{align}
    \mathrm{H}(p) &= -\sum_{\yy \in \mathrm{supp}(p)} p(\yy) \log p(\yy) \label{eq:ent}\\
    &= -\mathbb{E}_{\yy \sim p}\left( \log p(\yy)\right)\label{eq:ent_exp}
\end{align}
\noindent In the case of language models, $\mathrm{supp}(p)$ is the set of all strings that can be generated from the model's vocabulary $\vocab$. As this is exponentially large in $|\vocab|$, directly computing $\mathrm{H}(p)$ is intractable. We can use its equivalence to \cref{eq:ent_exp}, however, to estimate $\mathrm{H}(p)$ with a simple Monte-Carlo estimator:
\begin{equation}
    \mathrm{\hat{H}}(p) = -\frac{1}{K} \sum_{k=1}^K \log p(\yy^{(k)})
\end{equation}
where we sample $\yy^{(k)} \sim p$ for $k=1, \ldots, K$. 

\cref{tab:generated_seq_analysis} shows results from UID-regularized models compared with the baseline.
The models trained with the variance and local consistency regularizers exhibit a preference for longer sequence length and higher lexical diversity. Additionally, the entropy estimates of these models are notably higher, which, following the principle of maximum entropy \cite{jaynes1957information},\footnote{ The principle of maximum entropy states that the probability distribution that best represents the current knowledge state is the one with the largest entropy.} can be seen as an additional advantage of UID-regularized models over their unregularized counterparts. 

\subsection{UID Behavior}\label{subsec:uid_behavior}
To take a closer look at how UID regularization affects language models, we examine the relationship between minimizing perplexity and UID behavior, where we quantify UID behavior as the variance of models' surprisals. 
We consider models trained on the English EuroParl dataset with the variance regularizer at strengths $\beta \in \{0.01$, $0.03$, $0.05$, $0.07$, $0.09\}$ and our baseline (which is equivalent to $\beta = 0$),
For further comparison, we also train a model with $\beta = -0.01$ to observe the effects of \emph{penalizing} UID behavior.
We report results on the EuroParl test set in \cref{fig:uid_behavior}. 

We observe that the model trained with a UID penalty (negative $\beta$) indeed exhibits worse perplexity and UID behavior (variance of surprisals) on the test set. 
And as we might expect, models trained with higher $\beta$ exhibit UID behavior more strongly, as our quantification is part of their training objective. 
Overall, from $\beta=0.01$ to $\beta=0.05$, both perplexity and UID behavior are positively correlated with $\beta$, but when we optimize too much for UID ($\beta \geq 0.07$), there is a trade-off in which model perplexity begins to increase. 

We also observe an intriguing phenomenon in \cref{fig:uid_behavior}.
Models that achieve similar perplexity can have substantially different UID behavior values on the test set. 
Specifically, the $\beta = 0$ and $\beta = 0.07$ models, which have almost the same perplexity, have variance of surprisals of 17.8 and 15.8---a difference of more than ten percent!
If such models with similar perplexity can have varying definitions of what constitutes good UID behavior, then prior work, which has drawn conclusions on UID based on surprisals computed by a single model \cite{aylettturk04,jaeger-levy-07,jain-etal-2018-uniform}, may need revisiting. 
As this direction is outside the scope of the present paper, we leave it as future work. 

\input{figures/uid_behavior_1}

\section{Discussion and Related Work}\label{sec:discussion-related-work}
We discussed how operationalizing UID for language modeling leads to better models in a wide variety of settings. 
These results both provide a new form of evidence for the UID hypothesis and build on prior work exploring UID in modern-day NLP models.

\paragraph{Evidence for the UID hypothesis.} 
Our work extends the body of psycholinguistic research on uniform information density, which has largely corroborated the UID hypothesis by providing evidence that variation in surprisal, as estimated by a language model, is minimized in natural language.
In addition to early studies that used this approach to find evidence for UID in syntactic reduction \cite{jaeger-levy-07}, morphosyntactic contractions \cite{frank2008speaking}, and prosodic structure \cite{aylettturk04}, the same line of reasoning has been used by more recent work exploring a variety of other linguistic properties.
These studies have found that word duration can be predicted by syntactic surprisal \cite{demberg-etal-2012-syntactic,moore2013syntactic}, construction probability \cite{kuperman2012effects}, informativity \cite{seyfarth2014word}, and contextual predictability \cite{jurafsky2001probabilistic,bell2003effects,gahl2004knowledge}.
They have also observed that word length is reflected by conceptual complexity \cite{lewis2016length}; word order choice can be predicted by processing cost \cite{bloem-2016-testing,sikos2017information}; phonological patterns can be shaped by word predictability \cite{hall2018role}; and UID computed at the sequence level predicts human preferences for syntactic alternatives of the same sentence.

Whereas the above prior work has used language modeling as a tool for measuring UID, our paper has explored the exact converse---we have asked whether UID, operationalized as a regularizer, can be used as a tool for training better language models.
We argue that if the UID hypothesis holds as a general principle, then we should be able to exploit it as a training criterion that improves language modeling.
And accordingly, our results show that---across a variety of languages and dataset sizes---regularization for UID did indeed improve perplexity, which we view as an alternative kind of evidence for the UID hypothesis at scale.

Notably, \cref{fig:uid_behavior} at first could appear to contradict the UID hypothesis, since models with better UID behavior did not always achieve better perplexity. 
We do not consider this as evidence against the UID hypothesis, however. 
Rather, we posit that when $\beta$ is too large, we may be optimizing for UID to the point of tending towards unnatural language---a perfectly uniform dispersion of information across an utterance may come at the cost of strange lexical choices.
In this light, such a trade-off should be somewhat expected.

\paragraph{UID in modern NLP.}
In addition to the traditional line of psycholinguistic work, there have also been more-recent studies on UID in the context of modern NLP, although this work is relatively sparse.
\citet{rubino-etal-2016-information} leverage information density encoded as surprisal at the word, part of speech, and syntax levels to help build a state-of-the-art model for mixed-domain translationese detection.
\citet{jain-etal-2018-uniform} incorporate UID measures across sentences into models designed to detect natural versus manipulated text.
Perhaps the work that is most related to ours, \citet{meister-etal-2020-beam}, leverages UID to explain why beam search is an effective decoding algorithm and uses operationalizations of UID during beam search to alleviate problems with decoding poorly calibrated machine translation models. 
Whereas \citet{meister-etal-2020-beam} focuses on decoding, our work shows the first evidence that UID can be operationalized to aid training.

\section{Conclusions}

In closing, we have proposed encoding uniform information density as a regularizer for training language models---a novel manner of incorporating an established psycholinguistic theory into modern statistical language modeling. 
In experiments on a range of languages and dataset sizes, UID regularization consistently improves perplexity over baselines. 
Our results suggest that UID is a valid inductive bias for improving the canonical maximum likelihood objective in language modeling, providing a new, alternative type of evidence that supports the UID hypothesis at scale.
Our work opens the door to future research directions such as using similar techniques to validate other psycholinguistic phenomena, applying UID regularization in conditional language generation tasks, and exploring how UID regularized models perform in downstream NLP applications.

\section*{Ethical Concerns}
Language models have various ethical, environmental, and financial concerns.
We cannot do justice to them here, but do see \citet{10.1145/3442188.3445922} for a pointer.
We do not foresee any additional ethical concerns with the contributions made in our work beyond those discussed in \citet{10.1145/3442188.3445922}.

\section*{Acknowledgements}
We thank Roger Levy for feedback in the middle stages of our work and Tiago Pimentel, David Reitter, Tal Linzen, and Slav Petrov for feedback on the manuscript.

\bibliographystyle{acl_natbib}
\bibliography{anthology,acl2021}

\clearpage 
\onecolumn

\appendix
\section{Appendix}\label{sec:appendix-section}
\paragraph{Datasets.}
\cref{tab:data_stats} shows the train, dev, and test set splits for the language modeling datasets we use.
\input{tables/data_stats}

\paragraph{Hyperparameters.}
\cref{tab:hyperparams} shows the optimized $\beta$ hyperparameter from a grid-search over $\beta \in \{0.006$, $0.008$, $0.01$, $0.02$, $0.03$, $0.04$, $0.05\}$ for both regularizers on all datasets we use. 
Notably, the best $\beta$ for variance ranged from 1$\times 10^{-2}$ to 5$\times 10^{-2}$, and the best $\beta$ for local consistency ranged from 6$\times 10^{-3}$ to 2$\times 10^{-2}$.
For use on a new dataset, we recommend starting with 1$\times 10^{-2}$, which we found almost always improved perplexity for both regularizers (on these datasets, at least).
\input{tables/hyperparams}

\end{document}

%% file: tables/multiple_languages.tex
\begingroup
\begin{table}[H]
    \centering
    \small
    \begin{tabular}{l l} \toprule
        \underline{\textsc{Language}} (\# train tokens) & \multicolumn{1}{c}{Perplexity}  \\
        \midrule
        \underline{\textsc{Czech}} (12.4M) \\
        Baseline (no UID)  & 47.47 \\
         + UID: variance & 47.24 ($\hspace{0.3mm}\downarrow\hspace{-0.7mm} 0.5\%$) \\
         + UID: local consistency & \textbf{47.08} ($\hspace{0.3mm}\downarrow\hspace{-0.7mm} 0.8\%$)$^{\dagger}$ \\
        \midrule
        \underline{\textsc{English}} (46.7M) \\
        Baseline (no UID)  & 21.34  \\
         + UID: variance & \textbf{21.08} ($\hspace{0.3mm}\downarrow\hspace{-0.7mm} 1.2\%$)$^{\dagger}$ \\
         + UID: local consistency & 21.19 ($\hspace{0.3mm}\downarrow\hspace{-0.7mm} 0.7\%$)$^{\dagger}$ \\
        \midrule
        \underline{\textsc{Finnish}} (59.3M) \\
        Baseline (no UID) & 51.58 \\
         + UID: variance & \textbf{51.30} ($\hspace{0.3mm}\downarrow\hspace{-0.7mm} 0.5\%$)$^{\dagger}$ \\
         + UID: local consistency & 51.49 ($\hspace{0.3mm}\downarrow\hspace{-0.7mm} 0.2\%$) \\
        \midrule
        \underline{\textsc{French}} (51.3M) \\
        Baseline (no UID)  & 17.08 \\
         + UID: variance & \textbf{17.02} ($\hspace{0.3mm}\downarrow\hspace{-0.7mm} 0.4\%$)$^{\dagger}$ \\
         + UID: local consistency & 17.03 ($\hspace{0.3mm}\downarrow\hspace{-0.7mm} 0.3\%$)$^{\dagger}$ \\
        \midrule
        \underline{\textsc{German}} (42.3M) \\
        Baseline (no UID) & 26.62 \\
         + UID: variance & 26.50 ($\hspace{0.3mm}\downarrow\hspace{-0.7mm} 0.4\%$)$^{\dagger}$ \\
         + UID: local consistency & \textbf{26.45} ($\hspace{0.3mm}\downarrow\hspace{-0.7mm} 0.6\%$)$^{\dagger}$ \\
        \midrule
        \underline{\textsc{Indonesian}} (45.7M) \\
        Baseline (no UID)  & 53.96  \\
         + UID: variance & \textbf{53.66} ($\hspace{0.3mm}\downarrow\hspace{-0.7mm} 0.6\%$)$^{\dagger}$  \\
         + UID: local consistency & 53.70 ($\hspace{0.3mm}\downarrow\hspace{-0.7mm} 0.5\%$)   \\
        \midrule
        \underline{\textsc{Spanish}} (47.2M) \\
        Baseline (no UID)  & 22.54 \\
         + UID: variance & \textbf{22.37} ($\hspace{0.3mm}\downarrow\hspace{-0.7mm} 0.8\%$)$^{\dagger}$ \\
         + UID: local consistency & 22.44 ($\hspace{0.3mm}\downarrow\hspace{-0.7mm} 0.4\%$)$^{\dagger}$ \\
        \midrule
        \underline{\textsc{Swahili}} (6.3M) \\
        Baseline (no UID)  & 40.45  \\
         + UID: variance & 39.79 ($\hspace{0.3mm}\downarrow\hspace{-0.7mm} 1.6\%$)$^{\dagger}$  \\
         + UID: local consistency & \textbf{39.44} ($\hspace{0.3mm}\downarrow\hspace{-0.7mm} 2.5\%$)$^{\dagger}$ \\ 
        \midrule
        \underline{\textsc{Tagalog}} (4.2M) \\
        Baseline (no UID)  & 80.48  \\
         + UID: variance & 78.40 ($\hspace{0.3mm}\downarrow\hspace{-0.7mm} 2.5\%$)$^{\dagger}$  \\
         + UID: local consistency & \textbf{78.12} ($\hspace{0.3mm}\downarrow\hspace{-0.7mm} 2.9\%$)$^{\dagger}$   \\
        \midrule
        \underline{\textsc{Turkish}} (38.1M) \\
        Baseline (no UID)  & 66.13 \\
         + UID: variance & \textbf{65.70} ($\hspace{0.3mm}\downarrow\hspace{-0.7mm} 0.7\%$)$^{\dagger}$ \\
         + UID: local consistency & 66.06 ($\hspace{0.3mm}\downarrow\hspace{-0.7mm} 0.1\%$) \\
         \bottomrule
    \end{tabular}
    \caption{
    UID regularizers improve perplexity for multiple languages. 
    $^{\dagger}$ indicates statistical significance compared with the baseline ($p<0.05$).
    }
    \label{tab:multiple_languages}
\end{table}
\endgroup
\noindent

%% file: tables/dataset_size_1.tex
\begingroup
\setlength{\tabcolsep}{3.4pt}
\begin{table}[t]
    \centering
    \small
    \begin{tabular}{l l l l}
        \toprule 
         & WMT'06 & EuroParl & WT-103 \\
        \midrule
        \# training tokens & 16.0M & 47.0M & 103.2M \\
        \midrule
        Baseline (no UID) & 49.70 & 21.34 & 29.89 \\
         + UID: variance & \textbf{48.25}$^{\dagger}$ & \textbf{21.08}$^{\dagger}$ & \textbf{29.58} \\
         + UID: local consistency & 48.79 & 21.19 & 29.73 \\
         \bottomrule
    \end{tabular}
    \caption{
    UID regularizers improve perplexity on language models trained on English datasets of varying size. 
    Improvements tend to be larger on smaller datasets. 
    $^{\dagger}$ indicates statistical significance compared with the baseline ($p<0.05$).
    }
    \label{tab:dataset_size_1}
\end{table}
\endgroup

%% file: figures/dataset_size_2.tex
\pgfplotsset{width=6cm,height=5.0cm,compat=1.9}
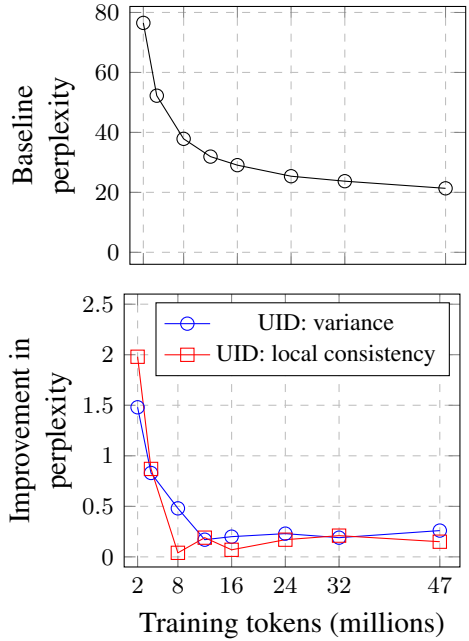
\begin{figure}[t]
\begin{centering}
\begin{tikzpicture}
\begin{axis}[
    ylabel={Baseline\\perplexity},
    xmin=0, xmax=50,
    ymin=-4, ymax=82,
    xtick={2, 8, 16, 24, 32, 47},
    ytick={0, 20, 40, 60, 80},
    legend pos=north east,
    ymajorgrids=true,
    xmajorgrids=true,
    ylabel style={align=center},
    grid style=dashed,
    xticklabels={,,}
]
\addplot[
    color=black,
    mark=o,
    mark size=2.5pt,
    ]
    coordinates {
    (2,   76.52)
    (4,   52.24)
    (8,   37.85)
    (12,  31.92)
    (16,  29.07)
    (24,  25.38)
    (32,  23.72)
    (47,  21.34)
    };
\end{axis}
\end{tikzpicture}
\pgfplotsset{width=6cm,height=5.2cm,compat=1.9}
\begin{tikzpicture}
\begin{axis}[
    xlabel={Training tokens (millions)},
    ylabel={Improvement in\\perplexity},
    xmin=0, xmax=50,
    ymin=-0.1, ymax=2.6,
    xtick={2, 8, 16, 24, 32, 47},
    ytick={0, 0.5, 1, 1.5, 2, 2.5},
    legend pos=north east,
    ymajorgrids=true,
    xmajorgrids=true,
    ylabel style={align=center},
    grid style=dashed,
    y label style={at={(axis description cs:-0.125,0.5)},anchor=south},
]
\addplot[
    color=blue,
    mark=o,
    mark size=2.5pt,
    ]
    coordinates {
    (2,   1.48)
    (4,   0.83)
    (8,   0.48)
    (12,  0.17)
    (16,  0.2)
    (24,  0.23)
    (32,  0.19)
    (47,  0.26)
    };
    \addlegendentry{UID: variance}
\addplot[
    color=red,
    mark=square,
    mark size=2.5pt,
    ]
    coordinates {
    (2,   1.98)
    (4,   0.87)
    (8,   0.04)
    (12,  0.19)
    (16,  0.07)
    (24,  0.17)
    (32,  0.21)
    (47,  0.15)
    };
    \addlegendentry{UID: local consistency}
\end{axis}
\end{tikzpicture}
    \vspace{-2mm}
\caption{
Improvement in perplexity for UID regularized models trained on subsets of varying size sampled from the EuroParl English dataset (full dataset size 47.0 million tokens).
UID regularization helped more when training data was more limited.
}
\label{fig:dataset_size}
\end{centering}
\end{figure}

%% file: tables/generated_text_evaluation.tex
\begingroup
\begin{table*}[t]
    \centering
    \small
    \begin{tabular}{l c c c c c}
        \toprule
         & Sequence & Model & \multicolumn{3}{c}{\% unique $n$-grams} \\
         & length & entropy & $n=2$ & $n=3$ & $n=4$ \\
        \midrule
        Baseline (no UID) & 22.9 & 69.6 & 37.7 & 73.5 & 90.9 \\
        + UID: variance & 24.0 & 79.4 & 40.7 & 77.8 & 93.3 \\
        + UID: local consistency & 23.3 & 73.9 & 39.1 & 75.7 & 92.1 \\
        \bottomrule
    \end{tabular}
    \caption{
    Text generated by UID-regularized language models is longer (higher average sequence length), higher entropy (computed via monte-carlo estimation), and more lexically diverse (a higher ratio of unique $n$-grams).  
    }
    \label{tab:generated_seq_analysis}
\end{table*}
\endgroup

%% file: tables/label_smoothing.tex
\begingroup
\begin{table}[t]
    \centering
    \small
    \begin{tabular}{l c c}
        \toprule
         & WMT'06 & EuroParl \\
        \midrule
        \# training tokens & 16.0M & 47.0M \\
        \midrule
        Baseline & 35.75 & 23.22 \\
         + label smoothing, $\alpha = 0.01$ & 36.15 & 26.26 \\
         + label smoothing, $\alpha = 0.05$  & 55.56 & 40.79 \\
         + label smoothing, $\alpha = 0.1$  & 90.57 & 68.26 \\
         \bottomrule
    \end{tabular}
    \caption{
    Label smoothing, another form of regularization that similarly augments the cross-entropy objective with a penalty, does not improve perplexity. (Results shown on dev set). 
    }
    \label{tab:label_smoothing}
\end{table}
\endgroup

%% file: figures/uid_behavior_1.tex
\pgfplotsset{width=7.2cm,height=7.0cm,compat=1.9}
\begin{figure}[t]
\begin{centering}
\begin{tikzpicture}
\begin{axis}[
    xlabel={Perplexity},
    ylabel={UID behavior\\(variance of surprisals)},
    xmin=20.95, xmax=22.05,
    ymin=14.85, ymax=18.65,
    xtick={21.0, 21.2, 21.4, 21.6, 21.8, 22.0},
    ytick={15.0, 15.5, 16, 16.5, 17, 17.5, 18, 18.5},
    legend pos=north east,
    ymajorgrids=true,
    ylabel style={align=center},
    xmajorgrids=true,
    grid style=dashed,
    nodes near coords, 
    nodes near coords align={center},
    every node near coord/.append style={font=\small, text=black},
    point meta=explicit symbolic,
]
\addplot[
    color=green,
    mark=*,
    mark size=2pt,
    nodes near coords style={xshift=5.1ex},
    ]
    coordinates {
    (21.48,   18.3)[$\beta = -0.01$]
    };
\addplot[
    color=blue,
    mark=oplus*,
    mark size=2pt,
    nodes near coords style={xshift=7ex},
    ]
    coordinates {
    (21.34,   17.82)[$\beta = 0$ (baseline)]
    };
\addplot[
    color=black,
    mark=*,
    mark size=2pt,
    nodes near coords style={xshift=5.1ex},
    ]
    coordinates {
    (21.15,   17.51)[$\beta = 0.01$]
    };
\addplot[
    color=black,
    mark=pentagon*,
    mark size=3pt,
    nodes near coords style={xshift=5.1ex},
    ]
    coordinates {
    (21.12,   16.96)[$\beta = 0.03$]
    };
\addplot[
    color=black,
    mark=*,
    mark size=2pt,
    nodes near coords style={xshift=5.1ex},
    ]
    coordinates {
    (21.19,   16.24)[$\beta = 0.05$]
    };
\addplot[
    color=black,
    mark=*,
    mark size=2pt,
    nodes near coords style={xshift=5.1ex},
    ]
    coordinates {
    (21.38,   15.76)[$\beta = 0.07$]
    };
\addplot[
    color=black,
    mark=*,
    mark size=2pt,
    nodes near coords style={xshift=5.1ex},
    ]
    coordinates {
    (21.58,   15.42)[$\beta = 0.09$]
    };

\end{axis}
\end{tikzpicture}
\vspace{-2mm}
\caption{
A trade-off between perplexity ($x$-axis) and variance of surprisals (a measure of UID behavior; $y$-axis). The black pentagon indicates the $\beta$ that yielded the best perplexity ($\beta = 0.03$). 
}
\vspace{-3mm}
\label{fig:uid_behavior}
\end{centering}
\end{figure}
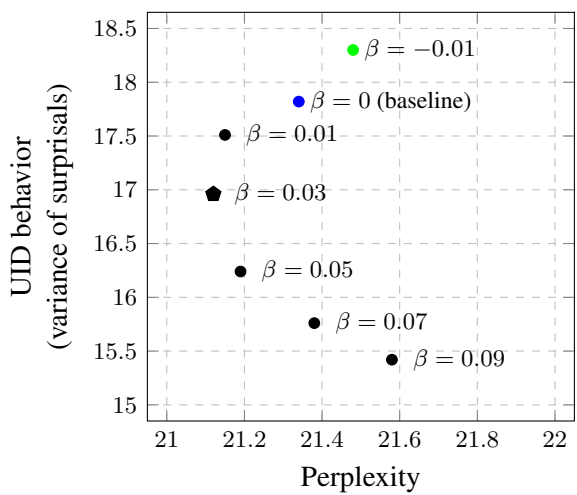

%% file: tables/data_stats.tex
\begingroup
\setlength{\tabcolsep}{3.1pt}
\begin{table*}[h]
    \centering
    \small
    \begin{tabular}{l l l l | r | r r | r r | r r}
        \toprule 
         & & & & Vocab & \multicolumn{2}{c}{Train} & \multicolumn{2}{|c|}{Dev} & \multicolumn{2}{c}{Test} \\
        Language & Family & Source & Split & size & Sentences & Tokens & Sentences & Tokens & Sentences & Tokens  \\
        \midrule
        English & Indo-European & EuroParl & 80--10--10 & 64k & 1.6M & 46.7M & 201k & 5.8M & 201k & 5.8M \\
         &  & WMT'06 & 80--10--10 & 62k & 751k & 17.0M & 2.0k & 61k & 3.1k & 90k \\
         &  & WT-103 & provided & 268k & 1.8M & 103.2M & 3.8k & 217k & 4.4k & 246k \\
        Czech & Indo-European & EuroParl & 80--10--10 & 64k & 517k & 12.4M & 65k & 1.6M & 65k & 1.6M \\
        French & Indo-European & EuroParl & 80--10--10 & 64k & 1.6M & 51.3M & 201k & 6.4M & 201k & 6.3M \\
        German & Indo-European & EuroParl & 80--10--10 & 64k & 1.5M & 42.3M & 192k & 5.4M & 192k & 5.2M \\
        Spanish & Indo-European & EuroParl & 80--10--10 & 64k & 1.6M & 47.2M & 197k & 6.0M & 197k & 5.9M \\
        Finnish & Uralic & Wiki-40B & provided & 128k & 256k & 59.3M & 14.1k & 3.9M & 14.0k & 3.2M \\
        Indonesian & Austronesian & Wiki-40B & provided & 128k & 156k & 45.7M & 8.7k & 3.1M & 8.6k & 2.5M \\
        Tagalog & Austronesian & Wiki-40B & provided & 128k & 26k & 4.2M & 1.5k & 270k & 1.4k & 220k \\
        Turkish & Turkic & Wiki-40B & provided & 128k & 143k & 38.1M & 7.8k & 2.5M & 7.7k & 1.9M \\
        Swahili & Niger-Congo & Wikipedia & 80--10--10 & 128k & 406k & 6.3M & 51k & 800k & 51k & 803k \\
        \bottomrule
    \end{tabular}
    \caption{
    Train, dev, and test splits, as well as vocab size, for the language modeling datasets that we use in this paper. If train-dev-test splits were provided, then we used them. Otherwise, we performed a 80--10--10 train-dev-test split. We found a vocab size of 64k to cover more than 98\% of the training set for the Indo-European languages, and a vocab size of 62k allowed us to cover 100\% in the training set of English WMT'06. 
    For the remaining languages, which had larger vocabularies, we followed Wiki-40B \cite{guo-etal-2020-wiki} and increased the vocab size to 128k.
    }
    \label{tab:data_stats}
\end{table*}
\endgroup

%% file: tables/hyperparams.tex
\begingroup
\begin{table*}[h]
    \centering
    \small
    \begin{tabular}{l l  c c | c c }
        \toprule
         & &  \multicolumn{4}{c}{UID Regularizer} \\
         & &  \multicolumn{2}{c}{Variance} & \multicolumn{2}{c}{Local Consistency} \\
        Language & Source & Best $\beta$ & Dev Loss & Best $\beta$ & Dev Loss \\
        \midrule
        English & EuroParl (full dataset) & 2$\times 10^{-2}$ & 4.519 & 8$\times 10^{-3}$ & 4.529 \\
         & EuroParl (2M subset) & 2$\times 10^{-2}$ & 6.497 & 1$\times 10^{-2}$ & 6.497 \\
         & EuroParl (4M subset) & 2$\times 10^{-2}$ & 5.940 & 1$\times 10^{-2}$ & 5.948 \\
         & EuroParl (8M subset) & 2$\times 10^{-2}$ & 5.500 & 8$\times 10^{-3}$ & 5.511 \\
         & EuroParl (12M subset) & 2$\times 10^{-2}$ & 5.236 & 8$\times 10^{-3}$ & 5.230 \\
         & EuroParl (16M subset) & 5$\times 10^{-2}$ & 5.084 & 2$\times 10^{-2}$ & 5.089 \\
         & EuroParl (24M subset) & 4$\times 10^{-2}$ & 4.841 & 2$\times 10^{-2}$ & 4.843 \\
         & EuroParl (32M subset)& 1$\times 10^{-2}$ & 4.747 & 1$\times 10^{-2}$ & 4.742 \\
         & WMT'06 & 3$\times 10^{-2}$ & 4.974 & 1$\times 10^{-2}$ & 4.991 \\
         & WT-103 & 1$\times 10^{-2}$ & 4.933 & 8$\times 10^{-3}$ & 4.939 \\
        Czech & EuroParl & 3$\times 10^{-2}$ & 5.388 & 1$\times 10^{-2}$ & 5.391 \\
        French & EuroParl & 1$\times 10^{-2}$ & 4.161 & 6$\times 10^{-3}$ & 4.162 \\
        German & EuroParl & 2$\times 10^{-2}$ & 4.782 & 8$\times 10^{-3}$ & 4.779 \\
        Spanish & EuroParl & 3$\times 10^{-2}$ & 4.539 & 1$\times 10^{-2}$ & 4.550 \\
        Finnish & Wiki-40B & 1$\times 10^{-2}$ & 5.811 & 6$\times 10^{-3}$ & 5.819 \\
        Indonesian & Wiki-40B & 3$\times 10^{-2}$ & 5.808 & 8$\times 10^{-3}$ & 5.809 \\
        Tagalog & Wiki-40B & 4$\times 10^{-2}$ & 6.319 & 8$\times 10^{-3}$ & 6.319 \\
        Turkish & Wiki-40B & 3$\times 10^{-2}$ & 6.119 & 8$\times 10^{-3}$ & 6.121 \\
        Swahili & Wikipedia & 2$\times 10^{-2}$ & 5.555 & 6$\times 10^{-3}$ & 5.546 \\
        \bottomrule
    \end{tabular}
    \caption{
    Best $\beta$ hyperparameters and dev losses for all experiments.
    }
    \label{tab:hyperparams}
\end{table*}
\endgroup